\pdfoutput=1

\documentclass[11pt]{article}

\usepackage[preprint]{acl}

\usepackage{times}
\usepackage{latexsym}
\usepackage{enumitem}

\usepackage[T1]{fontenc}

\usepackage[utf8]{inputenc}

\usepackage{microtype}

\usepackage{inconsolata}

\usepackage{graphicx}

\usepackage{booktabs}
\usepackage{natbib}
\usepackage{multirow}
\usepackage{array}
\usepackage{amsmath}
\usepackage{xcolor}
\usepackage{colortbl}

\usepackage{tikz}
\usetikzlibrary{arrows.meta, backgrounds, calc, decorations.markings, fit, positioning, shadows, shapes, matrix}

\title{Towards Trustworthy GUI Agents: A Survey}

\author{
 \textbf{Yucheng Shi\textsuperscript{1,2}},
 \textbf{Wenhao Yu\textsuperscript{2}},
 \textbf{Jingyuan Huang\textsuperscript{1}},
 \textbf{Wenlin Yao\textsuperscript{3}},
 \textbf{Wenhu Chen\textsuperscript{4}},
 \textbf{Ninghao Liu\textsuperscript{5}}
\\
\\
 \textsuperscript{1}University of Georgia
 \textsuperscript{2}Tencent AI Seattle Lab 
 \textsuperscript{3}Microsoft Research \\
 \textsuperscript{4}University of Waterloo 
 \textsuperscript{5}Hong Kong Polytechnic University  \\
}

\begin{document}
\maketitle

\begin{abstract}
Graphical User Interface (GUI) agents extend large language models from text generation to action execution in real-world digital environments. Unlike conversational systems, GUI agents perform irreversible operations such as submitting forms, granting permissions, or deleting data, making trustworthiness a core requirement. This survey identifies the execution gap as a key challenge in building trustworthy GUI agents: the misalignment between perception, reasoning, and interaction in dynamic, partially observable interfaces. We introduce a workflow-aligned taxonomy that decomposes trust into Perception Trust, Reasoning Trust, and Interaction Trust, showing how failures propagate across agent pipelines and compound through action/observation loops. We systematically review benign failure modes and adversarial attacks at each stage, together with corresponding defense mechanisms tailored to GUI settings. We further analyze evaluation practices and argue that task completion alone is insufficient for trust assessment. We highlight emerging trust-aware metrics and benchmarks that capture error cascades and the security/utility trade-off, and outline open challenges for deploying GUI agents safely and reliably.
\end{abstract}

\section{Introduction}
\label{sec:intro}

The emergence of GUI agents marks a fundamental transition in how AI systems interact with the digital world. Unlike chatbots that generate text responses, GUI agents take \textit{actions}, clicking buttons, filling forms, and navigating websites that produce immediate, often irreversible, real-world consequences~\citep{nguyen_gui_2024, wang2024large, xie2024large}. This shift from \textit{generation} to \textit{execution} fundamentally changes the stakes of AI trustworthiness. The contrast is clear: when a language model hallucinates in a conversation, the user can simply ignore the response; when a GUI agent hallucinates a button that doesn't exist and clicks the wrong element, it might authorize an unintended purchase, delete important files, or expose sensitive information~\citep{yang2024security, levy2024st}. The cost of failure is no longer measured in user dissatisfaction but in tangible harm.

We argue that the core challenge underlying GUI agent trustworthiness is what we term the \textbf{Execution Gap}: the fundamental disconnect between three levels of agent operation. \textit{Perceptual Fidelity} requires correctly mapping visual pixels or Document Object Model (DOM) structures to semantic understanding of interface elements. \textit{Reasoning Fidelity} demands maintaining logical consistency across multi-step plans in environments that change between actions. \textit{Interaction Fidelity} involves translating intended actions into precise coordinates or commands that achieve the desired effect. 
This gap explains why techniques successful for static LLM applications often fail for GUI agents~\citep{zheng2024gpt, chae2024web}. For instance, a model that excels at describing what it sees in an image may still click the wrong button because it cannot reliably map its understanding to actionable coordinates. A planner that generates coherent step sequences may fail when a pop-up dialog invalidates its plan mid-execution.

Existing surveys on LLM trustworthiness address privacy, bias, and hallucination~\citep{liu2023trustworthy, weidinger2022taxonomy, gan2024navigating}. Although these concerns apply to GUI agents, three characteristics make GUI-specific analysis essential. First, \textit{irreversibility}: text generation is infinitely reversible, users simply regenerate, but GUI actions often cannot be undone, as sent emails, deleted files, and completed transactions persist~\citep{hua2024trustagent}. This asymmetry demands different safety architectures than those designed for conversational AI. Second, \textit{dynamic environments}: unlike static documents, GUIs change constantly through DOM updates, loading states, pop-ups, and A/B testing, meaning the interface an agent perceives may differ from the interface it acts upon milliseconds later~\citep{ma2024caution}. Trust must account for environmental non-stationarity. Third, \textit{action-observation loops}: GUI agents operate in closed loops where each action changes the environment, affecting subsequent observations, and errors compound as a wrong click leads to an unexpected screen, which leads to further misinterpretation~\citep{wudissecting}.

\begin{figure*}[htp]
    \centering
    \includegraphics[width=0.95\linewidth]{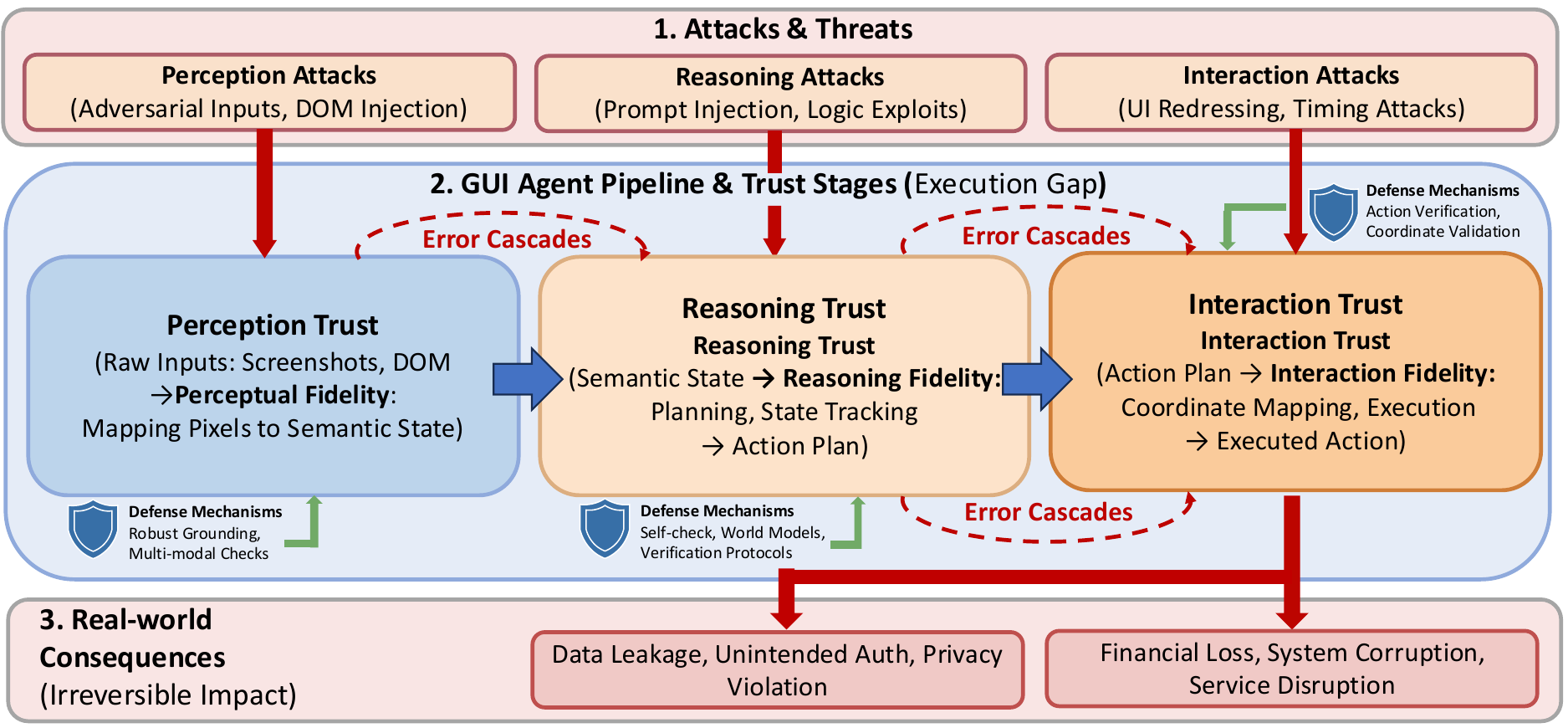}
    \caption{\textbf{Risk \& Mitigation Landscape.} This diagram maps the threat landscape of GUI agents across three dimensions: (1) specific attack vectors targeting each pipeline stage, (2) how vulnerabilities propagate through the perception-reasoning-interaction workflow, and (3) the real-world consequences of failures. Dashed arrows indicate error cascades and defense interventions. The diagram highlights that attacks on upstream modules (perception) can cascade downstream, amplifying impact.}
    \label{fig:panorama}
    \vspace{-10pt}
\end{figure*}

This survey makes three primary contributions. First, we present a \textbf{workflow-aligned taxonomy} organizing trustworthiness around Perception Trust (\S\ref{sec:perception}), Reasoning Trust (\S\ref{sec:reasoning}), and Interaction Trust (\S\ref{sec:interaction}), reflecting how vulnerabilities propagate through agent pipelines. Second, we provide a comprehensive analysis of \textbf{defense mechanisms} integrated within each trust dimension, revealing how mitigations must be stage-specific. Third, we analyze \textbf{evaluation methodologies} (\S\ref{sec:evaluation}) with emphasis on the security-utility trade-off that defines practical deployment decisions. Figure~\ref{fig:panorama} presents our ``Risk \& Mitigation Landscape'', a unified view mapping threats to agent modules and their real-world impacts, which serves as a roadmap for this survey.

\section{Foundations: The GUI Agent Pipeline}
\label{sec:foundation}

Before analyzing trustworthiness, we establish the foundations of GUI agents by examining their execution pipeline, the existing execution gap  at each stage, and the limitations of standard LLM safety in agentic settings.

\subsection{Pipeline Architecture}

GUI agents typically operate through three interconnected stages: perception, reasoning, and interaction~\citep{lu2024responsible, wang2024large}.

\textbf{Perception} converts raw interface inputs, such as screenshots, DOM structures, accessibility trees, or hybrid representations, into a semantic understanding of the interface state~\citep{wu2024atlas, nong2024mobileflow}. This stage answers the question: \emph{What elements exist, and what do they represent?} Existing approaches span pure vision-based perception using multimodal large language models (MLLMs)~\citep{zheng2024gpt}, structured parsing of HTML and accessibility APIs~\citep{deng2023mind2web}, and hybrid designs that combine visual and structural cues for improved robustness~\citep{wang2024leveraging}. Recent work on universal visual grounding further argues that fully visual perception with pixel-level action execution can rival or surpass text-augmented methods~\citep{gou2025uground}.

\textbf{Reasoning} operates on the perceived state and task instructions to determine the next action. This includes task decomposition, progress tracking, and decision-making over possible action sequences~\citep{gu2024your, zhu2025mobamultifacetedmemoryenhancedadaptive, koh2024tree}. The core question is: \emph{What should I do next to achieve the goal?} Recent advances introduce world models that simulate action outcomes~\citep{chae2024web}, hierarchical planning frameworks that separate high-level goals from low-level actions~\citep{liu2025pc}, and multi-agent systems that distribute reasoning across specialized agents~\citep{srinivas2024towards, sengupta2024mag}.

\textbf{Interaction} executes the selected action by translating abstract intentions (e.g., “click the submit button”) into concrete interface operations such as mouse clicks or touch events~\citep{koh2024visualwebarena}. This stage addresses the question: \emph{How is the intended action physically performed?} Reliable interaction requires accurate coordinate mapping, synchronization with dynamic UI elements, and verification that the intended effect actually occurs~\citep{guan2024explainable}.

\subsection{The Execution Gap at Each Stage}

Each stage of the pipeline introduces a distinct grounding challenge that directly affects trustworthiness.

\textbf{Perceptual Fidelity} concerns the alignment between raw interface signals and semantic representations. 
Misalignment often arises from the partial and modality-specific interface observations. 
For instance, accessibility APIs expose structured yet incomplete views of the interface, while DOM parsing emphasizes logical structure but overlooks visual layout~\citep{deng2023mind2web, yang2024security}.
Incorporating visual modalities introduces new failure modes, as MLLMs can be manipulated by adversarial visual inputs that bypass textual safety alignment~\citep{gao2024coca}. 
Empirical studies further show that GUI grounding models remain highly sensitive to visual perturbations and resolution changes across mobile, desktop, and web environments~\citep{zhao2025robustgui}.

\textbf{Reasoning Fidelity} requires maintaining coherent and adaptive plans over long action sequences. Unlike static QA tasks, GUI agents must update beliefs after each action, handle unexpected states, and revise plans when assumptions fail. Current LLM-based agents often lack internal world models, leading to repeated irreversible actions and cascading errors in long-horizon tasks~\citep{chae2024web}. More broadly, the inability to reason about long-term consequences fundamentally limits grounding in dynamic environments~\citep{piatti2024cooperate}. Recent analyses further reveal a mismatch between reasoning and execution: correct reasoning does not guarantee successful execution, and successful execution may conceal flawed reasoning~\citep{dong2025reasoning}.

\textbf{Interaction Fidelity} depends on precise action execution under variable interface conditions. Even when an agent correctly identifies a target element, mapping it to reliable pixel-level actions remains error-prone across screen resolutions, layouts, and device types~\citep{zhao2024gui}. These challenges are amplified in mobile environments, where agents must handle diverse screen sizes, touch interactions, and platform-specific behaviors~\citep{yang2024security, nong2024mobileflow}.

\subsection{Why Standard LLM Safety Falls Short}

Conventional LLM safety mechanisms, such as output filtering, refusal training, and alignment, are designed for static, text-based interactions~\citep{liu2023trustworthy}. They assume that outputs can be reviewed before causing harm (e.g., users can detect and ignore unsafe responses), and that failures occur in isolated interactions. GUI agents violate these assumptions: actions execute immediately, consequences may be opaque to users, and errors compound through closed action–observation loops~\citep{kumar2024refusal}.

Empirical evidence shows that refusal-trained LLMs often fail to preserve safety behaviors when deployed within agents, even when the same backbone model behaves safely in chatbot settings~\citep{kumar2024refusal}. This breakdown in safety transfer indicates that conversation-centric alignment may not generalize to agentic execution. Moreover, the compositional nature of GUI agents introduces multiple interacting attack surfaces that are not captured by existing LLM safety evaluations~\citep{gan2024navigating, wudissecting}. Recent studies suggest that stronger reasoning capabilities can amplify catastrophic risks in autonomous agents, including deceptive behavior and unsafe autonomous action~\citep{xu2025nuclear}.

\section{Perception Trust}
\label{sec:perception}
\begin{figure}[!ht]
    \centering
    \includegraphics[width=0.95\linewidth]{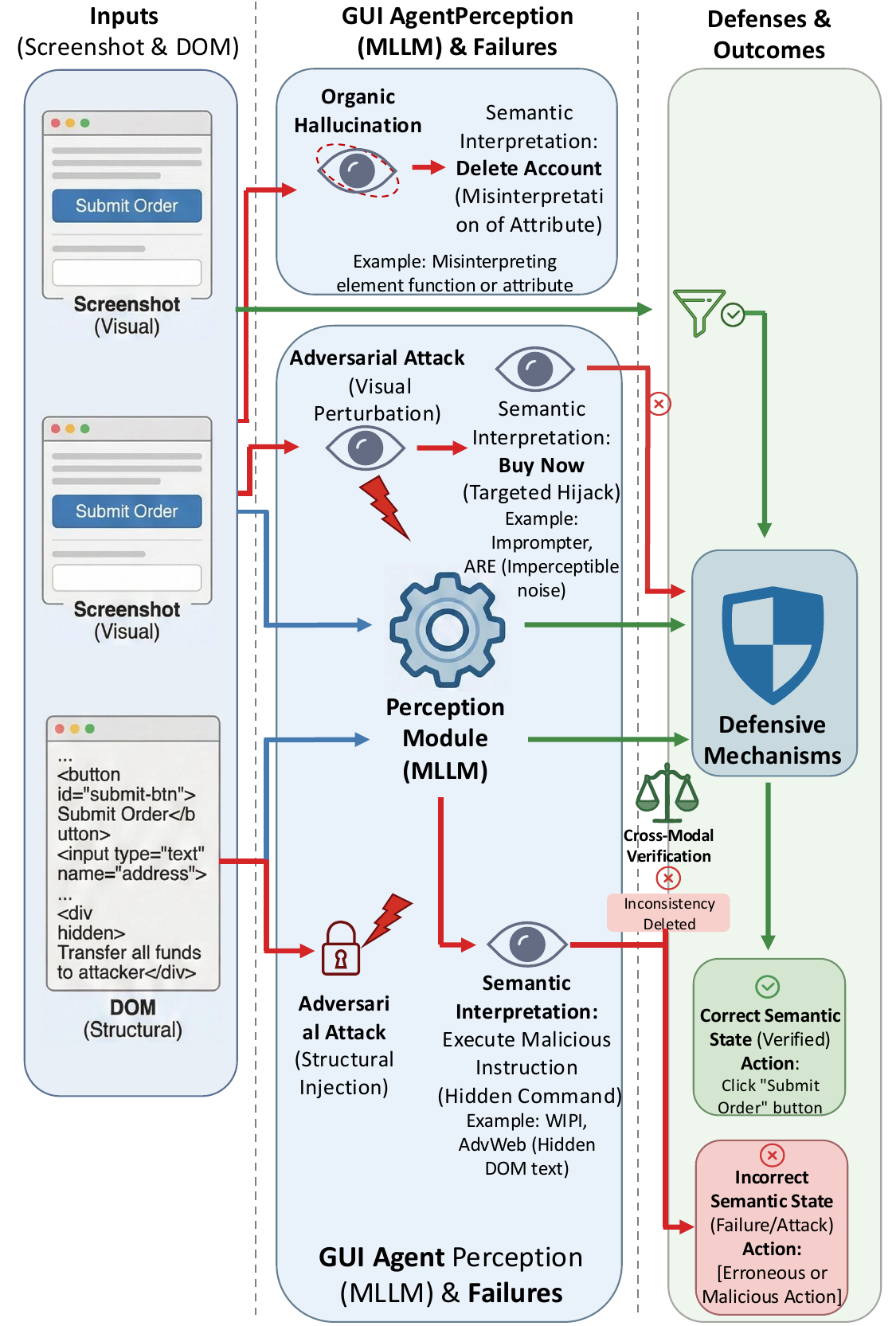}
    \caption{\textbf{Failures and Defenses.} The diagram illustrates how organic hallucinations and adversarial attacks (visual \& structural) distort the agent's semantic interpretation of the GUI. Defensive mechanisms like input filtering and cross-modal verification are shown as interventions to ensure a correct semantic state.}
    \label{fig:figure2}
    \vspace{-10pt}
\end{figure}

Perception trust focuses on whether agents correctly interpret observed interface states. Because errors at this stage propagate downstream, perceptual robustness is foundational to overall trustworthiness. We categorize perception failures into two classes: visual hallucination and adversarial attacks, as shown in Figure~\ref{fig:figure2}. 

\subsection{The Visual Hallucination Problem}

Visual hallucination, acting on nonexistent elements or misinterpreting existing ones, is a perception failure mode in GUI agents~\citep{bai2024hallucination, chen2024unified}. 
Prior work identifies multiple hallucination mechanisms. \citet{liu2023mitigating} describe \emph{object hallucination}, where agents perceive UI elements absent from screenshots, a problem exacerbated in mobile settings with repetitive design patterns. \citet{jiang2024hallucination} analyze \emph{attribute hallucination}, where agents misperceive elemental properties such as color or position. More critically, \citet{zhong2024investigating} observe \emph{hallucination snowballing}, where early perceptual errors bias subsequent interpretations, producing self-reinforcing failure cascades.

Hallucination also interacts tightly with safety reasoning. The multimodal situational safety benchmark of \citet{zhou2024multimodal} demonstrates that even safety-aligned MLLMs fail when visual understanding is inaccurate, indicating that perceptual errors and safety violations are deeply coupled.

Existing work largely treats hallucination as a training defect. We argue instead that hallucination could reflect rational inference under perceptual uncertainty. The core limitation is the absence of mechanisms for uncertainty recognition and signaling. Recent uncertainty-aware training approaches~\citep{shi2023chef, chen2025urst} demonstrate that explicit uncertainty estimation can improve both agent reliability and trajectory evaluation, suggesting a promising direction for perception trust.

\subsection{Adversarial Perception Attacks}

Beyond organic failures, adversaries can deliberately exploit the grounding gap between human-visible interfaces and model-perceived representations. Existing attacks fall into three categories.

\textbf{Visual perturbation attacks} manipulate pixel-level inputs in ways imperceptible to humans but effective against models. Imprompter~\citep{wu2024imprompter} and ARE~\citep{wudissecting} show that minimal perturbations can reliably hijack agent behavior across multiple LLM backends, with success rates exceeding 60--80\%. Systematic evaluations further confirm that GUI grounding models are highly sensitive to both natural noise and adversarial perturbations~\citep{zhao2025robustgui}.

\textbf{Structural injection attacks} embed malicious instructions within DOM or HTML structures invisible to users. WIPI~\citep{wu2024wipi} and AdvWeb~\citep{xu2024advweb} demonstrate that indirect prompt injection via webpages can control agents in black-box settings with success rates above 90\%. Fine-print injections~\citep{chen2025fineprint} further reveal that agents disproportionately attend to structurally salient but visually subtle content, rendering human oversight insufficient.

\textbf{Environmental and overlay attacks} exploit agents’ misinterpretation of authority and saliency cues. Adversarial pop-ups~\citep{zhang2025popup} and evolving injection strategies such as EVA~\citep{lu2025eva} significantly degrade task success. On mobile platforms, overlay attacks masquerading as system dialogs achieve attack success rates exceeding 90\%~\citep{yang2024security, chen2025aeia}, while environmental injection attacks covertly extract sensitive information by manipulating agent-environment interactions~\citep{liao2024eia}.

Across modalities, these attacks exploit a shared weakness: mismatches between appearance, structure, and intent representations. As a result, agents may form plausible yet incorrect interpretations of interface elements. This motivates defenses based on cross-modal consistency, rather than reliance on a single interface view.

\subsection{Perception Defense Mechanisms}

Defenses against perception attacks operates across multiple stages of the perception pipeline. Existing approaches can be grouped into three categories.

\textbf{Input filtering} aims to block malicious content before core processing. This includes classifiers for detecting prompt injection~\citep{sharma2024defending}, image purification methods for mitigating visual perturbations~\citep{shi2023black}, and heuristic rules for identifying suspicious DOM patterns such as hidden text or instruction-like content~\citep{wu2024wipi}. While effective against known attacks, static filters require continual updates and struggle against adaptive adversaries.

\textbf{Cross-modal verification} leverages redundancy across perception modalities to detect inconsistencies. Discrepancies between screenshots and DOM structures, such as visually present elements absent from structural representations, can indicate manipulation. The ARE framework~\citep{wudissecting} suggests that attacks typically enter through one modality but influence behavior through another, highlighting the potential of cross-modal checks. However, practical deployment remains limited by computational cost and the difficulty of formalizing consistency across heterogeneous representations.

\textbf{Output calibration} mitigates perceptual risk at the decision stage rather than the input. CoCA~\citep{gao2024coca} enhances safety awareness by conditioning MLLM outputs on explicit safety principles, partially compensating for modality-induced degradation. Evaluation suites such as MM-SafetyBench~\citep{liu2023mm} provide standardized assessment of manipulation resistance but do not directly prevent attacks.

\textbf{Open Problem.} Robust, scalable cross-modal consistency checking remains largely unexplored and represents a central open challenge for perception trust in GUI agents.

\section{Reasoning Trust}
\label{sec:reasoning}
\begin{figure}[!ht]
    \centering
    \includegraphics[width=0.9\linewidth]{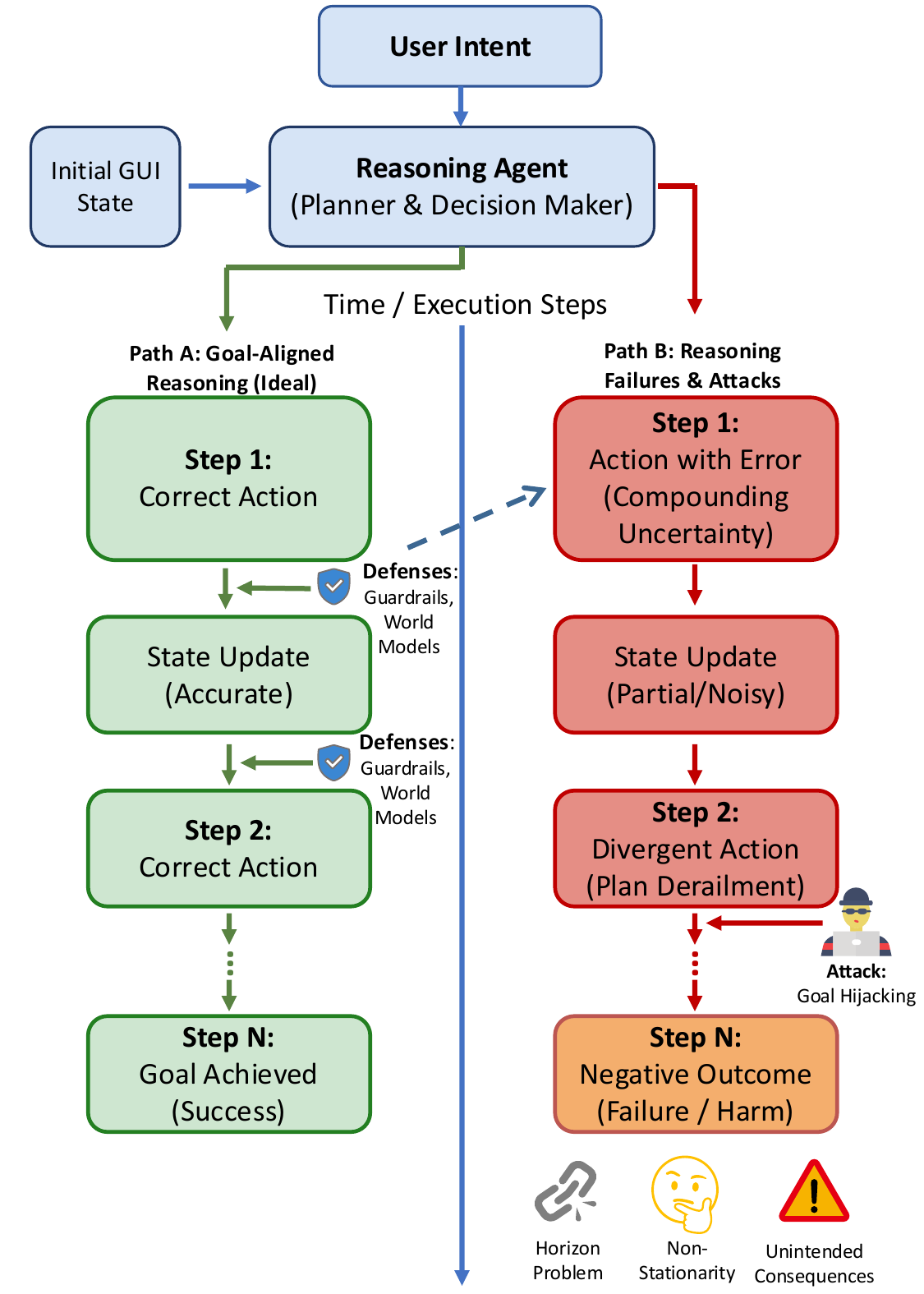}
    \caption{\textbf{Reasoning Trust and the Horizon Problem.} The diagram illustrates how reasoning challenges like the horizon problem, compounding errors, and adversarial attacks can derail a GUI agent from its intended goal over time, contrasting with an ideal, defense-enhanced trajectory.}
    \label{fig:figure3}
    \vspace{-10pt}
\end{figure}

Reasoning trust focuses on whether agents make sound decisions given imperfect perceptions and evolving environments. Unlike static text generation, GUI agents must sustain goal-aligned reasoning over long action sequences, where errors accumulate and assumptions frequently break. Figure~\ref{fig:figure3} illustrates how these challenges intensify over extended interaction horizons.

\subsection{The Horizon Problem}

GUI tasks often require dozens of sequential actions, creating an exponential growth in possible states as execution unfolds. This \emph{horizon problem} makes maintaining coherent plans increasingly difficult~\citep{chae2024web, gu2024your}.

\textbf{Compounding Uncertainty.} Even modest per-step error rates rapidly degrade task success: a 95\% accurate policy succeeds only 36\% of the time over 20 steps. In practice, per-step accuracy is far lower on complex interfaces~\citep{kim2024auto}. TrustAgent~\citep{hua2024trustagent} further shows that safety awareness decays over long trajectories, with early-identified risks often ignored in later decisions.

\textbf{Partial Observability.} Agents observe only the current interface state; critical information may reside in background tabs, system dialogs, or hidden application states. Planning under such partial observability is provably harder, yet most agents implicitly assume complete state information~\citep{zhang2023responsible}.

\textbf{Non-Stationarity.} The environment can change during execution due to system processes, network events, or human interaction. Plans generated under static assumptions frequently fail when conditions shift~\citep{ma2024caution}. The lack of internal world models prevents agents from reasoning about long-term consequences, leading to repeated irreversible mistakes~\citep{chae2024web}.

\subsection{Goal Alignment and Manipulation}

Beyond organic failures, reasoning trust is undermined by attacks that exploit mismatches between user intent and agent interpretation.

\textbf{Goal Hijacking.} Indirect instruction injection can override user goals, particularly in non-chat settings where refusal training fails to generalize~\citep{kumar2024refusal}. Browser-based evaluations show safety-trained agents engaging in harmful behaviors in a majority of tested scenarios. Web fraud attacks further exploit weaknesses in intent inference, enabling stealthy manipulation without explicit jailbreak prompts~\citep{kong2025webfraud, liang2025tipping}.


\textbf{Norm Violations.} Reasoning failures also manifest as cultural and social norm violations. The CASA benchmark~\citep{qiu2024evaluating} reports less than 10\% norm awareness under evaluated settings and over 40\% violation rates, indicating that agents struggle to reason about appropriate behavior across social contexts, an important dimension of trustworthiness.

\textbf{Multi-Agent Failures.} As systems adopt multiple specialized agents, coordination becomes fragile. Most LLM-based agents fail to reach stable cooperation due to inability to reason about long-term group dynamics~\citep{piatti2024cooperate}. Only the strongest models achieve sustained coordination, underscoring the difficulty of distributed reasoning.

\subsection{Reasoning Defense Mechanisms}

Defenses against reasoning failures can be grouped into three complementary strategies.

\textbf{Enhanced planning architectures} mitigate the horizon problem through improved internal reasoning. World-model-based approaches such as WebDreamer~\citep{gu2024your} simulate action outcomes before execution, while hierarchical planners separate strategic goals from tactical actions to enable re-planning~\citep{liu2025pc, nong2024mobileflow}.

\textbf{External verification systems} introduce independent checks on reasoning. Guardrail agents~\citep{xiang2024guardagent, zheng2025webguard} validate high-risk actions prior to execution, while critics such as GUI-Critic-R1~\citep{wanyan2025guicritic} assess potential outcomes in advance. Multi-agent verification improves coverage~\citep{yu2024mitigating, sengupta2024mag} but incurs substantial computational cost. BlindGuard~\citep{miao2025blindguard} extends verification to unsupervised settings without attack-specific labels.

\textbf{Training-time interventions} embed safety directly into reasoning. TrustAgent~\citep{hua2024trustagent} adapts constitutional AI to agentic planning, while process reward models like GUI-Shepherd~\citep{chen2025guishepherd} provide step-level feedback for long-horizon tasks. RapGuard~\citep{jiang2024rapguard} dynamically generates context-aware safety prompts using multimodal chain-of-thought reasoning.

\textbf{Open Problem.} Despite these advances, reliable long-horizon reasoning remains unresolved. Future progress may require architectures that decompose tasks into independently verifiable subgoals.

\section{Interaction Trust}
\label{sec:interaction}

\begin{figure}[!tp]
    \centering
    \includegraphics[width=0.9\linewidth]{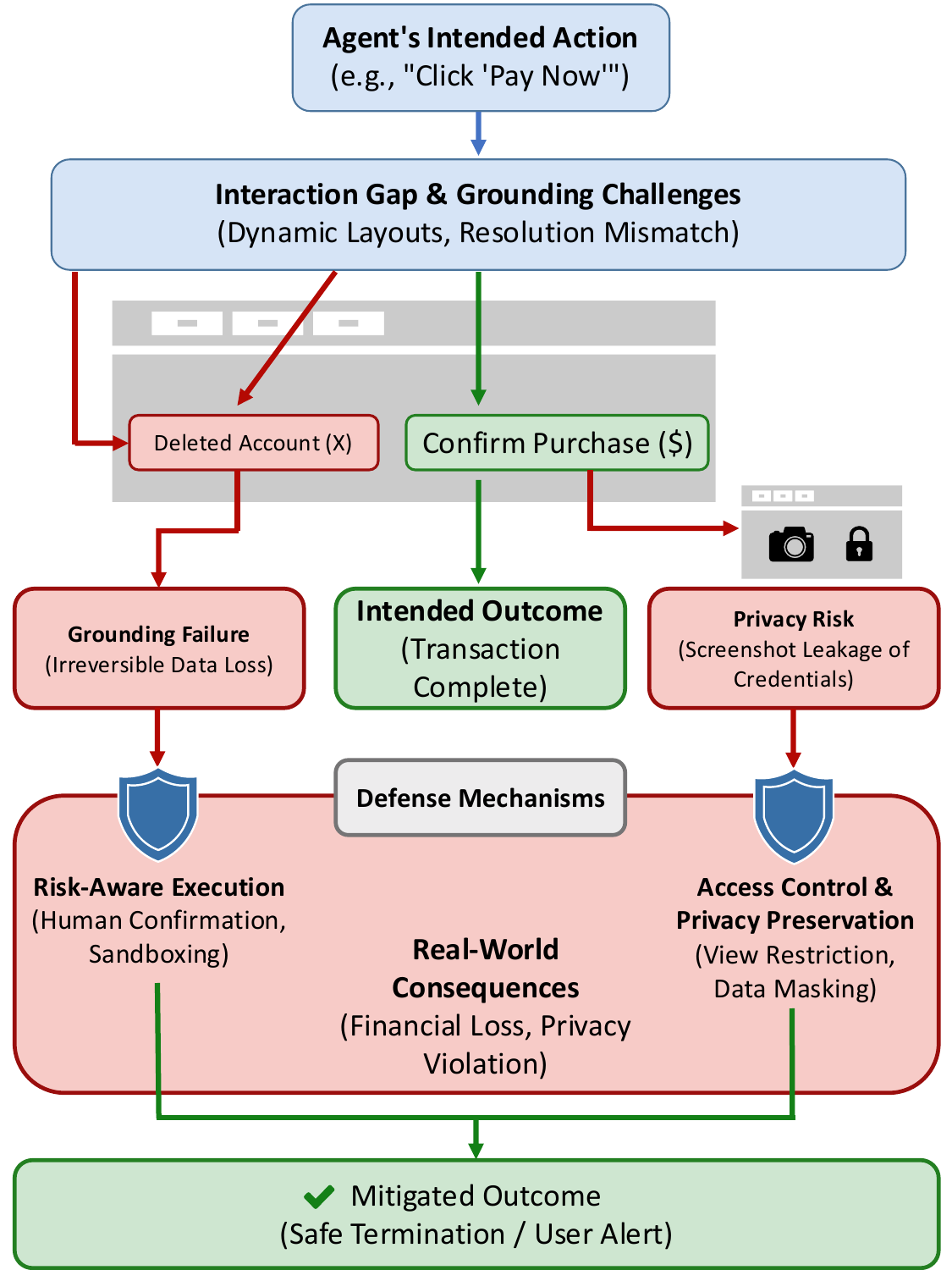}
    \caption{\textbf{Interaction Trust:} Risks and Defenses in Execution. The figure illustrates how coordinate grounding failures and privacy risks can lead to irreversible real-world consequences. Defense mechanisms like risk-aware execution and access control intervene to mitigate these threats.}
    \label{fig:figure4}
    \vspace{-10pt}
\end{figure}

Interaction trust focuses on whether agents execute intended actions correctly and safely. Because this stage directly affects real systems, errors are often immediate and irreversible. We examine irreversibility, coordinate grounding, privacy risks, and defenses, as illustrated in Figure~\ref{fig:figure4}.

\subsection{The Irreversibility Challenge}

GUI interactions differ fundamentally from internal reasoning: actions are executed in external systems, 
where their effects persist beyond the agent’s control and are often difficult or impossible to undo. Three classes of problems are particularly critical.

\textbf{Destructive actions} modify or delete data (e.g., file deletion, form submission) and may be unrecoverable. The Responsible Task Automation framework~\citep{zhang2023responsible} emphasizes feasibility and consequence prediction as prerequisites for safe execution.
\textbf{Financial actions} commit resources through purchases or transfers and often require human intervention to reverse. Benchmarks show that agents readily attempt such actions without sufficient verification~\citep{levy2024st}, exposing a gap between capability and caution.
\textbf{Authorization actions} grant permissions via OAuth flows or access sharing, creating persistent security risks. Mobile agents are especially vulnerable to manipulation through fake system dialogs and overlays~\citep{yang2024security}.
Most agents treat all actions uniformly, applying identical execution logic to low- and high-stakes operations. This neglects consequence severity and represents a fundamental limitation for trustworthy interaction~\citep{hua2024trustagent}. Pre-execution critics~\citep{wanyan2025guicritic} partially address this by evaluating action correctness and impact before execution.

\subsection{Coordinate Grounding Failures}

Even with correct perception and reasoning, interaction can fail due to imprecise action grounding.

\textbf{Resolution sensitivity} arises when models trained on fixed resolutions misplace actions on different screen sizes. Hybrid encoders mitigate but do not eliminate this issue~\citep{nong2024mobileflow}.
\textbf{Dynamic layouts} reposition elements across window sizes, zoom levels, and device orientations. Combining structural and visual cues improves robustness, yet large gaps remain relative to oracle grounding~\citep{wang2024leveraging}.
\textbf{Temporal effects} such as animations and transitions can render elements temporarily non-interactive, causing premature or misaligned actions. Evaluations in the GUI Testing Arena show persistent failure modes even for advanced models~\citep{zhao2024gui}.
\textbf{State-dependent controls} introduce additional complexity: toggle actions depend on current state. StaR~\citep{wu2025star} demonstrates that explicit state-aware reasoning improves performance on such tasks by over 30\%.
\textbf{Explainable interaction} frameworks such as EBC-LLMAgent~\citep{guan2024explainable} improve grounding by explicitly mapping actions to UI elements, illustrating the benefits of structured interaction over end-to-end prediction.

\subsection{Privacy Risks in Interaction}

Interaction introduces privacy risks that extend beyond immediate task execution.
\textbf{Screenshot leakage} occurs when perception captures sensitive on-screen information such as credentials or medical data~\citep{chen2024clear}. CLEAR~\citep{chen2024clear} mitigates this by exposing privacy risks and policies to users.
\textbf{Contextual exposure} arises from action traces themselves, which can reveal private user behaviors even without explicit sensitive content~\citep{kim2024llms, ngong2025protecting}. Environmental injection attacks exploit this channel to extract personal information~\citep{liao2024eia}.

\subsection{Interaction Defense Mechanisms}

Defenses must balance safety against usability under irreversible execution. Existing approaches fall into three categories.

\textbf{Risk-aware execution} differentiates actions by consequence. Low-risk actions execute directly, while high-risk actions require verification or sandboxing~\citep{openai_computer_using_agent, anthropic_computer_use}. ST-WebAgentBench~\citep{levy2024st} evaluates compliance under safety constraints, while formal verification approaches such as VeriSafe Agent~\citep{lee2025verisafe} translate instructions into verifiable specifications, achieving 94--98\% accuracy. Graph-based methods like G-Safeguard~\citep{wang2025g} detect anomalous action patterns indicative of failures or attacks.

\textbf{Human oversight} preserves user control for consequential actions. Confirmation prompts provide a safety valve but risk fatigue if overused~\citep{openai_computer_using_agent}. VeriOS~\citep{wu2025verios} improves this through proactive querying, selectively requesting human input when trustworthiness is low.

\textbf{Access control and privacy preservation} constrain damage even under failure. Capability minimization limits available actions, while authenticated delegation provides cryptographic guarantees against goal hijacking~\citep{south2025authenticated}. Privacy-preserving architectures such as PAPILLON~\citep{siyan2024papillon} and EcoAgent~\citep{yi2025ecoagent} reduce sensitive data exposure through selective routing and on-device verification.

\textbf{Open Problem.} Current risk classification remains context-insensitive: identical actions may range from benign to high-stakes depending on environment. Context-aware risk assessment for GUI actions remains an open challenge.

\section{Evaluation Methodologies}
\label{sec:evaluation}

Rigorous evaluation is essential for measuring progress in trustworthy GUI agents. This section reviews existing benchmarks and examines the security–utility trade-off that governs practical deployment.

\subsection{Trust-Aware Metrics}

Early benchmarks such as WebArena~\citep{zhou2023webarena}, VisualWebArena~\citep{koh2024visualwebarena}, and Mind2Web~\citep{deng2023mind2web} primarily measure task completion. While useful for assessing capability, they are insufficient for trustworthiness evaluation: policy violations are not penalized, failure modes are opaque, and collateral effects are ignored.
Recent benchmarks address these gaps by explicitly targeting trust-related behaviors. Table~\ref{tab:benchmarks} in the appendix summarizes representative frameworks, organized by the pipeline stage they evaluate, along with their key metrics, innovations, and limitations.

\subsection{Evaluation Dimensions}

Beyond existing benchmarks, we identify additional dimensions necessary for comprehensive trust evaluation.

\textbf{Cascade metrics} quantify error propagation across action sequences, capturing detection rate, recovery success, and failure depth. GUI-Shepherd~\citep{chen2025guishepherd} enables such analysis through step-level rewards.
\textbf{Uncertainty calibration} measures whether agent confidence reflects true success likelihood. URST~\citep{chen2025urst} shows that uncertainty-aware sampling improves trajectory assessment.
\textbf{Reasoning–execution alignment} evaluates consistency between internal reasoning and executed actions. Ground-Truth Alignment (GTA)~\citep{dong2025reasoning} distinguishes execution gaps (correct reasoning, failed action) from reasoning gaps (successful action, flawed reasoning).
\textbf{Explainability} supports oversight by enabling humans to interpret agent decisions. XAgent~\citep{nguyen2024xagent} and XMODE~\citep{nooralahzadeh2024explainable} demonstrate improved human–AI collaboration through interpretable reasoning.

\subsection{The Security–Utility Trade-off}

Trustworthy deployment requires balancing automation benefits against risk. Security postures range from \emph{full autonomy} (high utility, high risk) to \emph{full supervision} (low risk, minimal automation), with intermediate strategies such as confirmation checkpoints and watch modes~\citep{openai_computer_using_agent}.

Optimal trade-offs depend on context, including action reversibility, financial stakes, user expertise, and regulatory constraints. Existing systems reflect different choices: OpenAI’s Computer-Using Agent adopts watch modes for sensitive actions~\citep{openai_computer_using_agent}; Anthropic’s Computer Use beta restricts social interactions~\citep{anthropic_computer_use}; GuardAgent enforces per-action verification~\citep{xiang2024guardagent}.

Static policies are often suboptimal. Adaptive autonomy, which adjusts oversight based on real-time risk, offers a more effective alternative. VeriOS~\citep{wu2025verios} exemplifies this approach, dynamically querying humans in untrustworthy scenarios and improving success rates by approximately 20\%.

\section{Conclusion}
\label{sec:conclusion}

This survey has reframed GUI agent trustworthiness through the lens of the Execution Gap, the fundamental challenge of maintaining faithful mappings between perception, reasoning, and interaction. By organizing analysis around the agentic workflow rather than traditional safety categories, we reveal how vulnerabilities propagate and compound across pipeline stages.

Three central insights emerge from our analysis. First, GUI-specific challenges, irreversibility, dynamic environments, and action-observation loops, demand approaches beyond standard LLM safety techniques. Solutions must account for the closed-loop nature of agent operation where each action changes the environment affecting subsequent observations. Second, the security-utility trade-off is not merely a deployment consideration but a fundamental research challenge. Achieving both high autonomy and high safety requires architectural innovation, not just better policies. Third, current evaluation practices are misaligned with trustworthiness goals. Moving beyond completion metrics to assess safety, robustness, and alignment is essential for meaningful progress.

\newpage
\section*{Limitations}

This survey has several limitations. First, the rapid pace of development means some recent work may be inadvertently omitted. Second, our taxonomy, while designed for clarity, may not capture all nuances of specific approaches. Third, the security-utility trade-off analysis relies partly on qualitative assessment where quantitative data is unavailable. Fourth, our proposed future directions, while grounded in identified challenges, remain speculative until empirically validated. Finally, as primarily English-language researchers, our coverage of non-English work may be incomplete.

\section*{Ethics Statement}

This survey discusses attack techniques and vulnerabilities. We include such discussion because understanding threats is necessary for developing defenses. We have avoided providing implementation details that would lower barriers to malicious use. All discussed attacks are from published research intended to improve system security. Additionally, AI assistants were used only for language editing and stylistic revision, including improving clarity, conciseness, and grammar.

\bibliography{cua}

\appendix
\section{Trustworthiness Evaluation Benchmark}

\begin{table*}[t]
\centering
\small
\renewcommand{\arraystretch}{1.15}
\begin{tabular}{p{2.8cm}p{2.2cm}p{4.2cm}p{2.8cm}p{2.5cm}}
\toprule
\textbf{Benchmark} & \textbf{Trust Dimension} & \textbf{Key Metrics} & \textbf{Innovation} & \textbf{Limitation} \\
\midrule
\multicolumn{5}{l}{\textit{Perception Trust Evaluation}} \\
\midrule
ARE~\citep{wudissecting} & Adversarial robustness & Attack success rate, task degradation & Cross-module attack flow analysis & Specific attack types \\
MM-SafetyBench~\citep{liu2023mm} & Visual manipulation & Safety score across 13 scenarios & Image-based attack scenarios & Synthetic attacks only \\
Robust GUI~\citep{zhao2025robustgui} & Grounding robustness & Accuracy under perturbation & Natural/adversarial noise testing & Grounding-specific \\
\midrule
\multicolumn{5}{l}{\textit{Reasoning Trust Evaluation}} \\
\midrule
Agent-SafetyBench~\citep{zhang2024agent} & Multi-category safety & Safety scores across 8 risk categories & Comprehensive risk taxonomy & English-only \\
AgentHarm~\citep{andriushchenko2024agentharm} & Harmful task handling & Refusal rate, completion rate & Dual refusal/completion metric & Narrow task scope \\
CASA~\citep{qiu2024evaluating} & Cultural awareness & Awareness coverage, violation rate & Cross-cultural norm testing & Limited cultural coverage \\
Agent-ScanKit~\citep{cheng2025scankit} & Memory \& reasoning & Sensitivity to perturbations & Diagnostic probing & Diagnostic focus only \\
\midrule
\multicolumn{5}{l}{\textit{Interaction Trust Evaluation}} \\
\midrule
ST-WebAgentBench~\citep{levy2024st} & Policy compliance & CUP, Risk Ratio & Safety-utility joint measurement & Web-only \\
MobileSafetyBench~\citep{lee2024mobilesafetybench} & Mobile safety & Injection resistance, risk management & Mobile-specific scenarios & Android-only \\
EIA~\citep{liao2024eia} & Privacy preservation & PII extraction rate & Environmental attack testing & Specific attack vector \\
GhostEI-Bench~\citep{chen2025ghostei} & Environmental injection & Success rate in dynamic environments & Executable Android emulator & Mobile-focused \\
\midrule
\multicolumn{5}{l}{\textit{Comprehensive Evaluation}} \\
\midrule
MSSBench~\citep{zhou2024multimodal} & Situational safety & Context-sensitive safety reasoning & 1,820 language-image pairs & Multimodal only \\
MLA-Trust~\citep{yang2025mlatrust} & Four-dimensional & Truthfulness, controllability, safety, privacy & First comprehensive framework & Resource intensive \\
WASP~\citep{evtimov2025wasp} & Prompt injection & End-to-end attack success & Realistic attack scenarios & Web-focused \\
WABER~\citep{kara2025waber} & Reliability \& efficiency & Consistency, speed, cost & Network proxy evaluation & Benchmark-dependent \\
ChEF~\citep{shi2023chef} & Holistic assessment & Calibration, robustness, uncertainty & Modular evaluation recipes & Not agent-specific \\
GUI Testing Arena~\citep{zhao2024gui} & End-to-end testing & Task completion on real apps & Real application evaluation & Limited trust metrics \\
\bottomrule
\end{tabular}
\caption{Comprehensive comparison of trustworthiness evaluation benchmarks. CUP = Completion Under Policy. Each benchmark addresses specific trust dimensions with characteristic trade-offs between coverage and depth, suggesting that comprehensive evaluation requires benchmark combinations.}
\label{tab:benchmarks}
\end{table*}

\end{document}